\documentclass{article}

\usepackage{amsmath,amssymb,amsfonts}
\usepackage{algorithmic}
\usepackage{graphicx}
\usepackage{textcomp}
\usepackage{caption,subcaption}

\usepackage[utf8]{inputenc}
\usepackage[english]{babel}
\usepackage{csquotes}

\usepackage[symbol]{footmisc}
\renewcommand{\thefootnote}{\fnsymbol{footnote}}

\usepackage[style = ieee]{biblatex}
\newcommand\blfootnote[1]{%
  \begingroup
  \renewcommand\thefootnote{}\footnote{#1}%
  \addtocounter{footnote}{-1}%
  \endgroup
}

\usepackage{arxiv}

\usepackage[utf8]{inputenc} % allow utf-8 input
\usepackage[T1]{fontenc}    % use 8-bit T1 fonts
\usepackage{hyperref}       % hyperlinks
\usepackage{url}            % simple URL typesetting
\usepackage{booktabs}       % professional-quality tables
\usepackage{amsfonts}       % blackboard math symbols
\usepackage{nicefrac}       % compact symbols for 1/2, etc.
\usepackage{microtype}      % microtypography
\usepackage{lipsum}		% Can be removed after putting your text content
\usepackage{graphicx}
\usepackage{doi}

\usepackage{tikz}
\usepackage{textcomp}
\newcommand\copyrighttext{%
\footnotesize \textcopyright 2021 IEEE. Personal use of this material is permitted.
Permission from IEEE must be obtained for all other uses, in any current or future
media, including reprinting/republishing this material for advertising or promotional
purposes, creating new collective works, for resale or redistribution to servers or
lists, or reuse of any copyrighted component of this work in other works.

\doi{10.1109/ACCESS.2021.3051454}}
\newcommand\copyrightnotice{%
\begin{tikzpicture}[remember picture,overlay]
\node[anchor=south,yshift=20pt] at (current page.south) {\fbox{\parbox{\dimexpr\textwidth-\fboxsep-\fboxrule\relax}{\copyrighttext}}};
\end{tikzpicture}%
}

\title{Improved Static Hand Gesture Classification on Deep Convolutional Neural Networks using Novel Sterile Training Technique}

\author{ \href{https://orcid.org/0000-0002-3388-4805}{\includegraphics[scale=0.06]{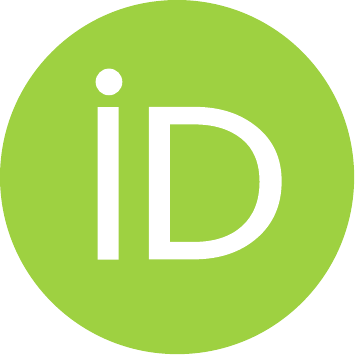}\hspace{1mm}Josiah W. Smith} \\
	Department of Electrical and Computer Engineering\\
	The University of Texas at Dallas\\
	Richardson, TX 75080 \\
	\texttt{josiah.smith@utdallas.edu} \\
	\And
	\href{https://orcid.org/0000-0002-0427-3773}{\includegraphics[scale=0.06]{orcid.pdf}\hspace{1mm}Shiva Thiagarjan} \\
	Department of Electrical and Computer Engineering\\
	The University of Texas at Dallas\\
	Richardson, TX 75080 \\
	\And
	\hspace{1mm}Richard Willis\thanks{Conducted this research while an undergraduate student at UTD ECE, but is currently affiliated with Citigroup Global Markets Inc., 390 Greenwich St. New York, NY 10013.} \\
	Department of Electrical and Computer Engineering\\
	The University of Texas at Dallas\\
	Richardson, TX 75080 \\
	\And
	\href{https://orcid.org/0000-0002-4322-0068}{\includegraphics[scale=0.06]{orcid.pdf}\hspace{1mm}Yiorgos Markis} \\
	Department of Electrical and Computer Engineering\\
	The University of Texas at Dallas\\
	Richardson, TX 75080 \\
	\texttt{yiorgos.makris@utdallas.edu} \\
	\And
	\href{https://orcid.org/0000-0001-7229-1765}{\includegraphics[scale=0.06]{orcid.pdf}\hspace{1mm}Murat Torlak} \\
	Department of Electrical and Computer Engineering\\
	The University of Texas at Dallas\\
	Richardson, TX 75080 \\
	\texttt{torlak@utdallas.edu} \\
}

\date{}

\renewcommand{\headeright}{\textit{IEEE Access} (Accepted)}
\renewcommand{\undertitle}{\textit{IEEE Access} (Accepted)}
\renewcommand{\shorttitle}{Improved Static Hand Gesture}

\hypersetup{
pdftitle={Improved Static Hand Gesture Classification on Deep Convolutional Neural Networks using Novel Sterile Training Technique},
pdfsubject={cs.CV, cs.AI, eess.SP},
pdfauthor={Josiah W.~Smith, Shiva Thiagarajan, Richard Willis, Yiorgos Makris, Murat Torlak},
pdfkeywords={Convolutional Neural Networks, Deep Learning, Hand Gesture Recognition, Millimeter-Wave Radar, Sterile Training},
}

\begin{document}
\maketitle
\copyrightnotice

\begin{abstract}
In this paper, we investigate novel data collection and training techniques towards improving classification accuracy of non-moving (static) hand gestures using a convolutional neural network (CNN) and frequency-modulated-continuous-wave (FMCW) millimeter-wave (mmWave) radars. 
Recently, non-contact hand pose and static gesture recognition have received considerable attention in many applications ranging from human-computer interaction (HCI), augmented/virtual reality (AR/VR), and even therapeutic range of motion for medical applications. 
While most current solutions rely on optical or depth cameras, these methods require ideal lighting and temperature conditions. 
mmWave radar devices have recently emerged as a promising alternative offering low-cost system-on-chip sensors whose output signals contain precise spatial information even in non-ideal imaging conditions. 
Additionally, deep convolutional neural networks have been employed extensively in image recognition by learning both feature extraction and classification simultaneously. 
However, little work has been done towards static gesture recognition using mmWave radars and CNNs due to the difficulty involved in extracting meaningful features from the radar return signal, and the results are inferior compared with dynamic gesture classification. 
This article presents an efficient data collection approach and a novel technique for deep CNN training by introducing ``sterile'' images which aid in distinguishing distinct features among the static gestures and subsequently improve the classification accuracy. 
Applying the proposed data collection and training methods yields an increase in classification rate of static hand gestures from $85\%$ to $93\%$ and $90\%$ to $95\%$ for range and range-angle profiles, respectively.
\end{abstract}

\blfootnote{This work was supported in part by Texas Instruments through the Foundational Technology Research Centre and the Texas Analog Center of Excellence.}

% keywords can be removed
\keywords{Convolutional Neural Networks \and Deep Learning \and Hand Gesture Recognition \and Millimeter-Wave Radar \and Sterile Training}

\section{Introduction}
\label{sec:introduction}
Accurately classifying human hand gestures has recently received significant attention as non-contact human-computer interaction (HCI) sensors become increasingly prevalent and desirable. Many efforts have been done towards classifying moving (dynamic) hand gestures and non-moving (static) hand gestures using optical cameras and many different classifiers \cite{baek2013comparison}. Applications of static gesture classification include augmented/virtual reality (AR/VR) \cite{son2017image}, human-computer interaction \cite{matilainen2016OUHANDS}, and even medical applications for range of motion and therapeutic applications \cite{anaz2018classification}. Such optical systems offer high-resolution two-dimensional (2-D) images but have innate drawbacks requiring specific lighting conditions and lacking depth information. Some solutions have investigated the use of an RGB-D depth camera \cite{lin2013_3D_hand_posture_RGBD}, but these devices suffer under sunlight, restricting their usage to indoors only \cite{son2017image}. On the other hand, small form-factor millimeter-wave (mmWave) frequency-modulated-continuous-wave (FMCW) radar offers high-resolution depth information but does not have the cross-range resolution of an optical camera. mmWave radars are advantageous over optical solutions, due to the semi-penetrative nature of the electromagnetic (EM) at the wavelengths in the mmWave frequency range and independence from ambient temperature effects, allowing for fine measurements in non-ideal lighting and temperature environments including occlusion, fog, indoor/outdoor, etc. Additionally, FMCW mmWave radar allows for simultaneous gesture classification and localization. High-resolution spatial information reflected from a human hand is embedded in the FMCW return signal. However, due to the nature of the FMCW radar as a time-of-flight (ToF) sensor and hardware size limitations, an off-the-shelf radar device cannot reconstruct an image reminiscent of a human hand, meaningful to the human eye. Thus, a deep convolutional neural network (CNN) approach is commonly adopted to classify dynamic gestures from radar return signals, after some preprocessing \cite{kim2017application}. Further, extracting meaningful features from the radar return signal is a key step towards accurately classifying hand gestures. As such, recent work on mmWave sensors for hand gesture recognition has been limited to dynamic hand gestures focusing on Doppler and micro-Doppler features \cite{dekker2017gesture,suh2018_24GHz_gesture,kim2016hand,zhang2016dynamic} with little attention being paid to the static gesture case \cite{kim2017staticgesture} due to the low classification rates in such applications. Prior research towards impulse-radio ultra-wideband (IR-UWB) \cite{leem2020detecting,park2016_ir_uwb_gesture,kim2017staticgesture} and Doppler radar-based gesture recognition \cite{dekker2017gesture,suh2018_24GHz_gesture,kim2016hand} using CNNs has shown promising results. However, with the exception of \cite{kim2017staticgesture}, gesture recognition on IR-UWB and Doppler radar has been limited to dynamic, moving gestures. IR-UWB and Doppler sensors are capable of classifying dynamic gestures with ease as motion is easily visible in the Doppler spectrum, whereas both these approaches suffer for static gestures as the distinct features to each gesture class are much more difficult to extract from a stationary human hand. Doppler radar employs the Doppler shift principle to provide relative velocity information of gestures but does not provide spatial features necessary for classifying stationary, static gestures. Therefore, for the problem of static gesture recognition, Doppler radar does not perform well as the hand remains stationary, thus making features of the hand reflection quite difficult to extract from a Doppler radar return signal. On the other hand, IR-UWB sensors act similarly to mmWave FMCW radars and must overcome the same inherent difficulty of classifying the higher-dimensional hand pose from the lower-dimensional radar return signal as discussed later. In this paper, we introduce novel techniques for data collection and CNN training intended to overcome the aforementioned limitations of mmWave sensors for static gesture recognition by providing distinct, meaningful features of each hand gesture thus aiding the CNN learning process.

The rest of this article is formatted as follows. In Section \ref{sec:fmcw_radar}, we briefly overview the FMCW radar signal model and key concepts helpful in developing an intuition for the static gesture classification problem. In Section \ref{sec:measurement_setup}, the measurement setup is discussed and the novelty of "sterile" data for FMCW radar is introduced. Section \ref{sec:dcnn} overviews some basics of deep convolutional neural networks and highlights the network architecture employed in our implementation. In Section \ref{sec:classification_results}, classification results are shown and discussed, followed finally by conclusions.

\section{FMCW Radar Signal Model}
\label{sec:fmcw_radar}
Understanding the FMCW radar signal model provides several key insights into the difficulty of the static hand gesture recognition problem space. The frequency-modulated-continuous-wave signal model is well explored in the literature \cite{smith2020nearfieldisar} and briefly overviewed here to provide some insight into the inherent challenges of static gesture recognition using mmWave FMCW sensors.

\subsection{FMCW Beat Signal}
\label{subsec:fmcw_beat_signal}
We will begin by considering a single bistatic FMCW transceiver, whose transmitter and receiver are positioned at the points ($0$,$y_T$,$Z_0$) and ($0$,$y_R$,$Z_0$) in ($x$,$y$,$z$) space, respectively, and one stationary ideal point reflector in the scene with reflectivity $\sigma$ located at the point ($x_0$,$y_0$,$z_0$). The radar transceiver is positioned on the $x'$-$y'$ plane located at $z = Z_0$ from the point target.

First, the FMCW device generates what is known as a chirp signal, which can be modeled as a complex sinusoidal whose frequency increases linearly with time as
\begin{equation}
    m(t) = e^{j2\pi(f_0t + \frac{1}{2}Kt^2)}, \quad 0 \leq t \leq T,
\end{equation}
where $f_0$ is the instantaneous frequency at the time $t=0$, $K$ is the chirp slope, and $T$ is the chirp duration in fast time. The chirp bandwidth can easily be computed using $B = KT$ \cite{yanik2019near}.

The chirp signal $m(t)$ is transmitted by the transmit antenna, reflected off of the ideal point reflector, and returned to the receive antenna as a scaled and time-delayed version of the transmitted signal. Taking round-trip amplitude decay into account, the received signal can be modeled as
\begin{equation}
    \hat{m}(t) = \sigma \frac{m(t-\tau)}{R_T R_R} = \frac{\sigma}{R_T R_R} e^{j2\pi(f_0(t-\tau) + \frac{1}{2}K(t-\tau)^2)},
\end{equation}
where $\tau$ is the round-trip time delay \cite{yanik2018millimeter} and the values $R_T$ and $R_R$ (see Fig. \ref{fig:hand_scenario}) can be computed by
\begin{gather}
    R_T = \sqrt{x_0^2 + (y_0-y_T)^2+(z_0-Z_0)^2}, \\
    R_R = \sqrt{x_0^2 + (y_0-y_R)^2+(z_0-Z_0)^2}.
\end{gather}

Therefore, the round trip time delay $\tau$ can be computed by
\begin{equation}
    \tau = \frac{R_T+R_R}{c},
\end{equation}
where $c$ is the speed of light.

Now, the received signal $\hat{m}(t)$ is demodulated with the transmitted signal $m(t)$ yielding what is known as the IF signal or FMCW beat signal, written as
\begin{equation}
\label{eq:fmcw_beat_signal}
    s(t) = \frac{\sigma}{R_T R_R}e^{j2\pi(f_0\tau +K\tau - \frac{1}{2}K\tau^2)}
\end{equation}
The last phase term of (\ref{eq:fmcw_beat_signal}) is called the residual video phase (RVP) term and is known to be negligible \cite{yanik2019sparse}. Finally, the beat signal can be simplified to the expression
\begin{equation}
\label{eq:fmcw_beat_signal_k_multistatic}
    s(y_T,y_R,k) = \frac{\sigma}{R_T R_R} e^{jk(R_T + R_R)},
\end{equation}
where $k = 2\pi f/c$ is the wavenumber corresponding to the instantaneous frequency $f = f_0 + Kt$ for $t \in [0,T]$.

\subsection{Multistatic-to-Monostatic Conversion}
\label{subsec:mult_to_mono}
The result in (\ref{eq:fmcw_beat_signal_k_multistatic}) shows the FMCW beat signal from a single point reflector using a multistatic antenna array where the transmitter and receiver are not co-located. To ease the subsequent signal processing, it is desirable to approximate this multistatic echo signal to a monostatic version. This approximation already has been explored in the literature and we will simply use the result derived in \cite{yanik2019sparse,yanik2019cascaded}. The multistatic-to-monostatic conversion is known as a simple phase adjustment applied to each transceiver pair as
\begin{equation}
\label{eq:mult-to-mono}
    \hat{s}(y',k) = s(y_T,y_R,k) e^{-jk\frac{d_y^2}{4Z_0}},
\end{equation}
where $d_y$ is the small separation between the transmitter and receiver and $Z_0$ is an approximate distance from the radar to the target.

Now, the monostatic approximation yields a beat signal whose virtual antenna positions are at the midpoint of each of the transceiver pairs. Taking $y'$ as these virtual antenna locations, the virtual monostatic signal can be approximated by a simplified version of (\ref{eq:fmcw_beat_signal_k_multistatic}) as
\begin{equation}
\label{eq:fmcw_beat_signal_k_monostatic}
    \hat{s}(t) \approx \frac{\sigma}{R_0^2} e^{j2kR_0},
\end{equation}
where $R_0$ is the distance between each virtual antenna element and the point reflector and is expressed as
\begin{equation}
    R_0 = \sqrt{x_0^2 + (y_0-y')^2 + (z_0-Z_0)^2}.
\end{equation}

Now, the range of the target is clearly embedded in the frequency of the beat signal.

\subsection{FMCW Range-Angle Analysis}
\label{subsec:fmcw_range_angle_analysis}
Considering the ideal point reflector described previously and a uniform linear monostatic array along the $y$-axis, the range and range-angle profiles can be computed and used to localize the point reflector. First, as evident in (\ref{eq:fmcw_beat_signal_k_monostatic}) and (\ref{eq:distrubted_target}), the frequency of the beat signal corresponds directly to the range of the target. Thus, the range profile can be generated by performing a fast-Fourier transform (FFT) along the $k$-domain of the beat signal. The minimum resolvable distance between two targets, or range resolution, can be easily computed as $\Delta z_{min} = c/(2B)$ \cite{sheen2010near}. 

Similarly, as discussed in \cite{TI:rao2017intro}, the angular profile of the target scene can be computed by performing an FFT across the spatial $y$ domain. To avoid aliasing in the angle domain, by Nyquist theorem, the maximum theoretical distance between elements is $\lambda/4$, where $\lambda$ is the wavelength. However, even after applying both range and angle FFTs, the resulting range-angle profile only describes the intensity in two dimensions and is limited by the number of antenna elements and bandwidth.

In this work, we will employ and compare both range and range-angle analysis to preprocess the echo signals before using them for CNN training or classification.

\subsection{Modeling a Distributed Target}
\label{subsec:distributed_target}
For gesture recognition, a human hand can be mathematically modeled as a distributed target consisting of continuously varying reflectivity across space. Understanding how the radar captures such target scenes provides an intuition into the difficulty of the hand gesture recognition problem.

\begin{figure}[h]
    \centering
    \includegraphics[width=0.6\textwidth]{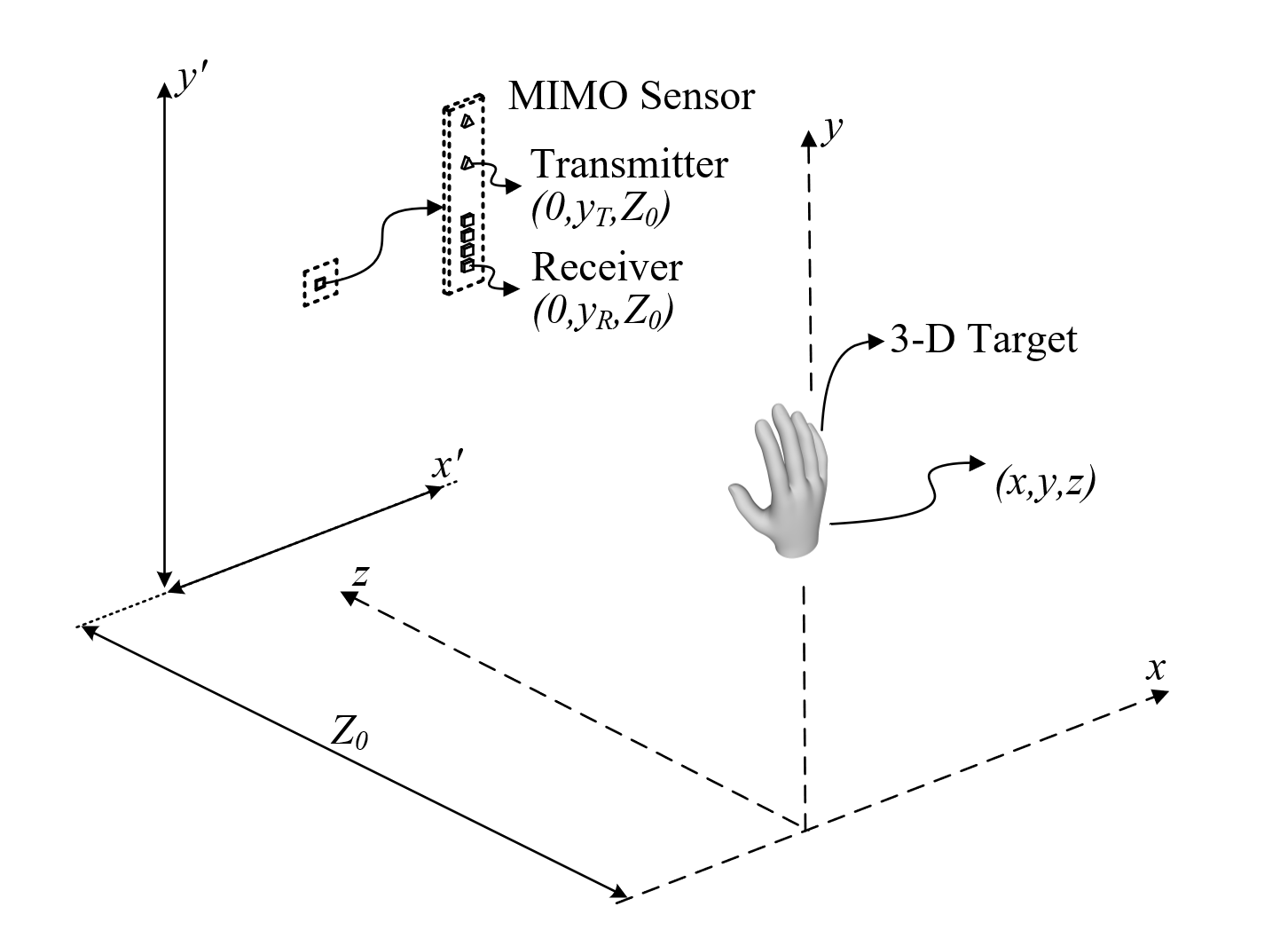}
    \caption{A MIMO radar sensor with transmitter and receiver antenna elements located at ($0$,$y_T$,$Z_0$) and ($0$,$y_R$,$Z_0$), respectively captures the return signal from a three-dimensional (3-D) target whose reflectivity function is $\sigma(x,y,z)$.}
    \label{fig:hand_scenario}
\end{figure}

Assuming a simple linear multistatic array along the $y$-axis, such as the depiction in Fig. \ref{fig:hand_scenario}, after the aforementioned conversion, the return signal from a distributed target can be modeled as the superposition of the echo signals from each of the target coordinates scaled by the target's reflectivity function $\sigma(x,y,z)$. The beat signal from each virtual monostatic transceiver at the positions $y'$ can be expressed as
\begin{equation}
\label{eq:distrubted_target}
    s(y',k) = \iiint \frac{\sigma(x,y,z)}{R^2}e^{j2kR}dxdydz.
\end{equation}
where $R$ is the radial distance from each virtual monostatic element located at the positions $y'$ to each point in the distributed target domain as 
\begin{equation}
    R = \sqrt{x^2 + (y-y')^2 + (z-Z_0)^2}.
\end{equation}

If samples are taken throughout the $x'$-$y'$ plane, the reflectivity function can be reconstructed by inverting (\ref{eq:distrubted_target}); however, for an application such as hand gesture recognition, the transceiver elements only span a small space along the $y'$-axis. This model provides insight into the simultaneous plausibility and difficulty of the static gesture recognition problem on FMCW radar. 

Embedded in the beat signal are high-resolution spatial features describing the shape of the target or static gesture being performed, meaning different hand poses or static gestures have distinct echo signals unique to that gesture. However, the target scene, or hand, cannot be analytically reconstructed as a three-dimensional (3-D) image and used to easily classify the gestures using traditional optical image approaches. Thus, classifying static hand gestures involves attempting to learn a 3-D pattern (the hand pose in three dimensions) from 2-D radar data. 

Further, another issue inherent to the hand gesture problem is the small radar cross-section (RCS) of the human hand resulting in a low signal-to-noise-ratio (SNR), as discussed later. Even with a large amount of data, since the RCS of the hand is low, the features unique to each gesture class are not pronounced. As a result, the CNN has difficulty discerning meaningful features for each gesture. Our proposed method using "sterile" data aims to overcome this deficiency in the training data and will be discussed in detail in the next section.

\begin{figure}[ht]
    \centering
    \includegraphics[width=0.75\textwidth]{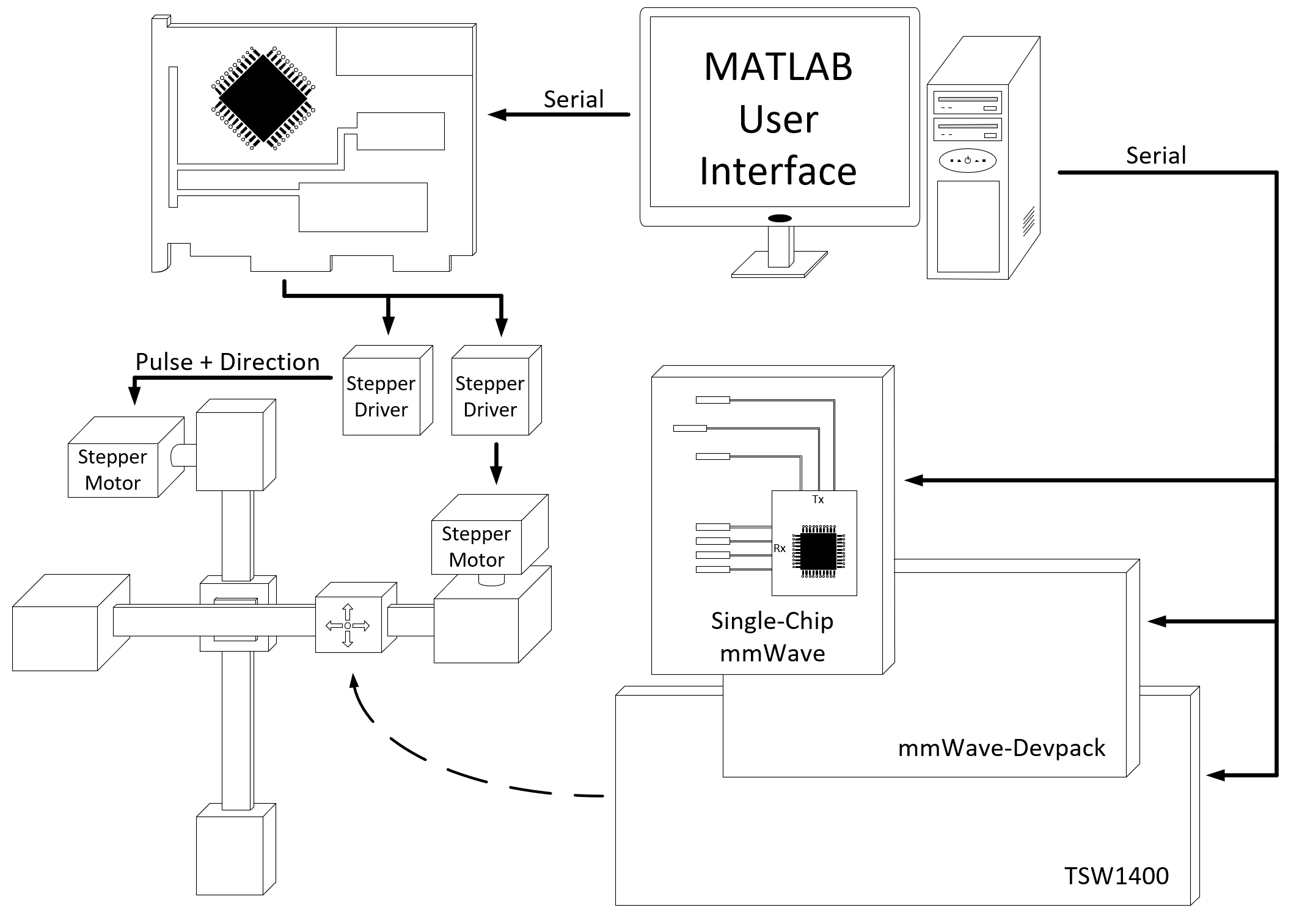}
    \caption{Two dimensional $x$-$y$ rectangular scanner system diagram.}
    \label{fig:xy_scanner_system_diagram}
\end{figure}

\section{Measurement Setup}
\label{sec:measurement_setup}
For any problem using the supervised learning approach to deep learning, the availability of meaningful and diverse data is crucial to building an accurate and well-generalized model. 

\subsection{Mechanical Scanner}
\label{subsec:mechanical_scanner}
To efficiently gather data from many perspectives, we first design a 2-D mechanical scanning system capable of positioning the radar anywhere within a $0.5$ m x $0.5$ m square, as shown in Fig. \ref{fig:xy_scanner_system_diagram}. The entire system is controlled by a custom-built MATLAB graphical user interface (GUI) hosted by a desktop computer. An AMC4030 motion controller receives commands over serial and controls the stepper motors via stepper drivers, accurately moving the radar to the desired location. Each stepper motor is mounted to a linear belt-driven rail with a usable length of $0.5$ m. The radar under test is an IWR1443BOOST, which is an off-the-shelf automotive radar from Texas Instruments (TI). As shown in the system diagram, the radar board is oriented with its MIMO array aligned vertically. The IWR1443BOOST has 3 transmit (Tx) antennas and 4 receive (Rx) antennas, but this work will exclusively use the 2 Tx and 4 Rx antennas which form a linear MIMO array whose virtual elements are separated by $\lambda/4$. The operating frequency is $77$ GHz and the bandwidth is $4$ GHz. The radar board is attached to a booster, the TI mmWave-Devpack, and a TI TSW1400 data capture card. All three radar boards are controlled via TI mmWave Studio, which receives commands from the MATLAB GUI. As such, the user has full control over the radar position, settings, timing, etc. directly from the custom GUI. 

\begin{figure}[h]
    \centering
    \includegraphics[width=0.35\textwidth]{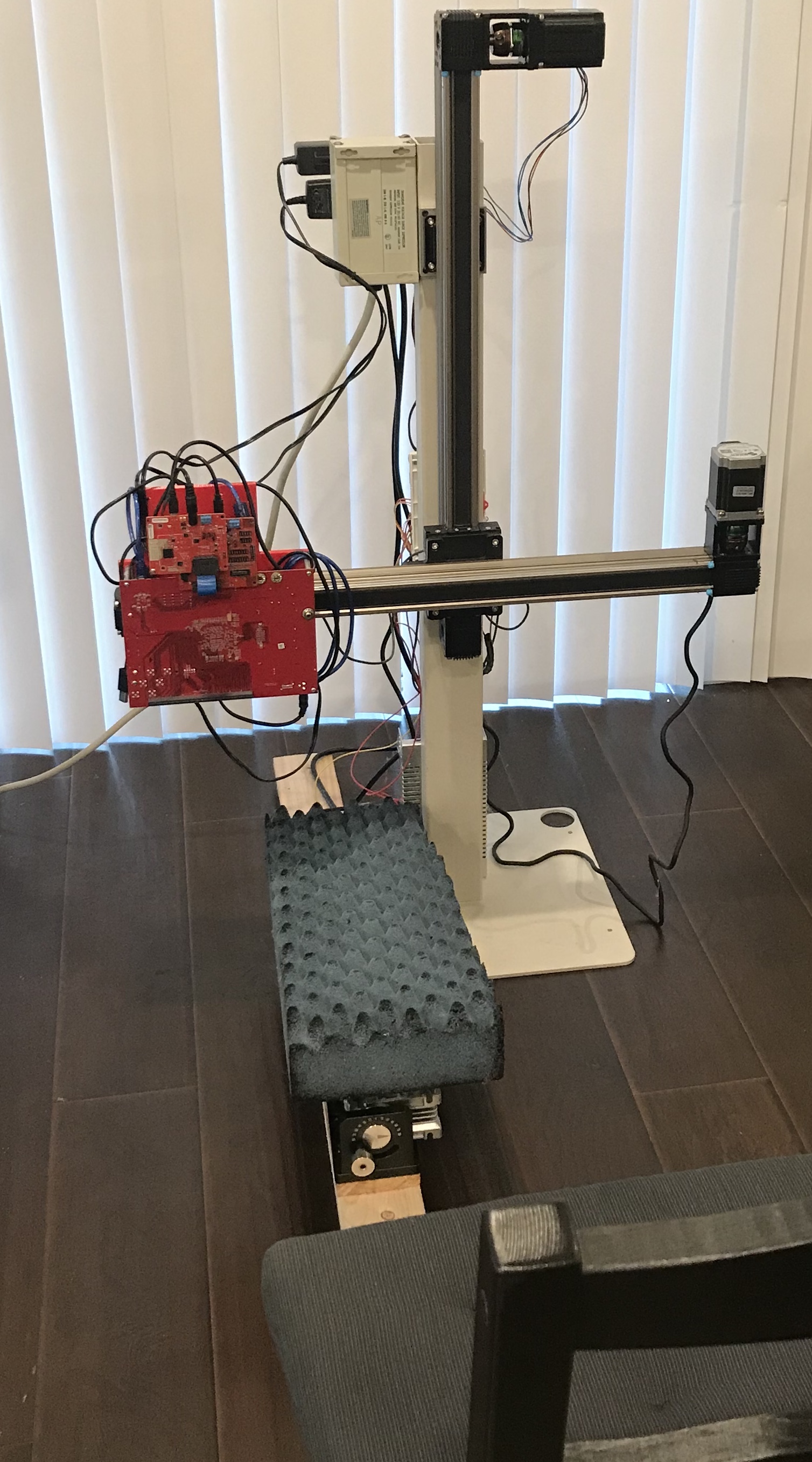}
    \caption{Two dimensional $x$-$y$ rectangular scanner system with chair for test subject to sit in.}
    \label{fig:xy_scanner_pic}
\end{figure}

The novel mechanical scanning system allows for the efficient capture of static hand gestures from many locations, as shown in Fig. \ref{fig:xy_scanner_pic}. A test subject simply keeps their hand in the correct pose and position and the radar is scanned both horizontally and vertically capturing many different perspectives of the hand gesture. 

For the rest of this article, the scanning range will be limited to a square with sides of $0.25$ m, and the subject's hand will be placed between $0.25$ m and $0.55$ m from the radar. The subject will hold their hand out in front of them performing the gesture with their hand kept away from their torso so as to avoid occluding the hand in the torso peak's side-lobes. Further, the user will sit in a chair located $1$ m from the radar while performing the gestures. Multiple test subjects are used to collect data varying in height, weight, torso size, arm length, hand size, and gender to diversify the dataset. Since the radar data are captured at locations throughout the $x'$-$y'$ plane, even though the subject does not move their hand, the dataset will consist of many unique "views" of the hand as if the hand is positioned at many locations relative to the radar. To the authors' knowledge, this article is the first attempt to use a 2-D scanner to collect static hand gesture data on mmWave radar. The mechanical scanner is employed to collect a diverse dataset consisting of multiple perspectives of the hand gestures; however, the problem of human hand reflectivity and feature prominence remain. The novel data collection technique is next extended in attempt to overcome these issues.

\subsection{Challenges}
\label{subsec:challenges}
\begin{figure}
     \centering
     \begin{subfigure}[b]{0.2\textwidth}
         \centering
         \includegraphics[width=\textwidth]{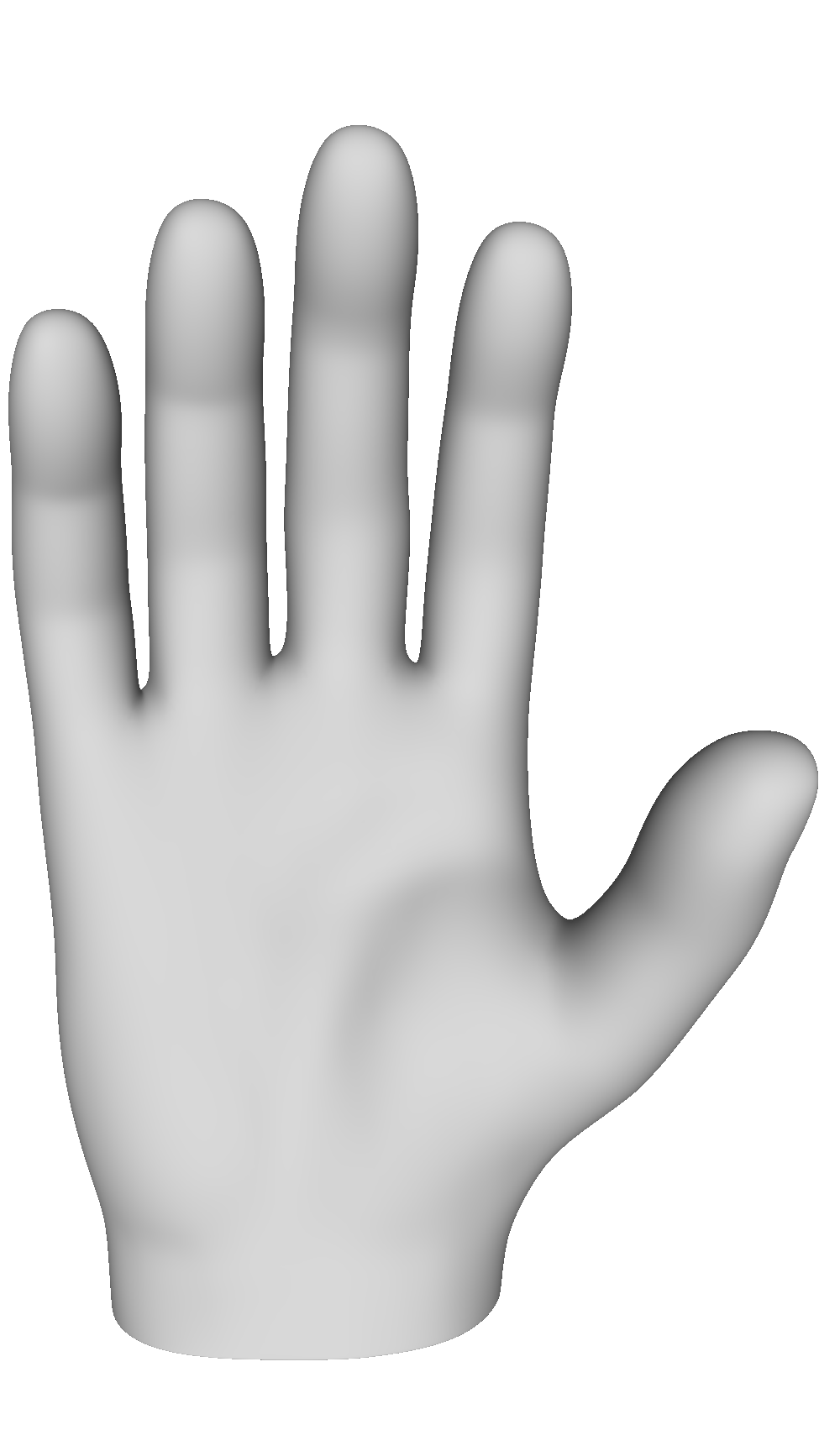}
         \caption{"Palm" hand gesture}
         \label{fig:palm}
     \end{subfigure}
     \begin{subfigure}[b]{0.2\textwidth}
         \centering
         \includegraphics[width=\textwidth]{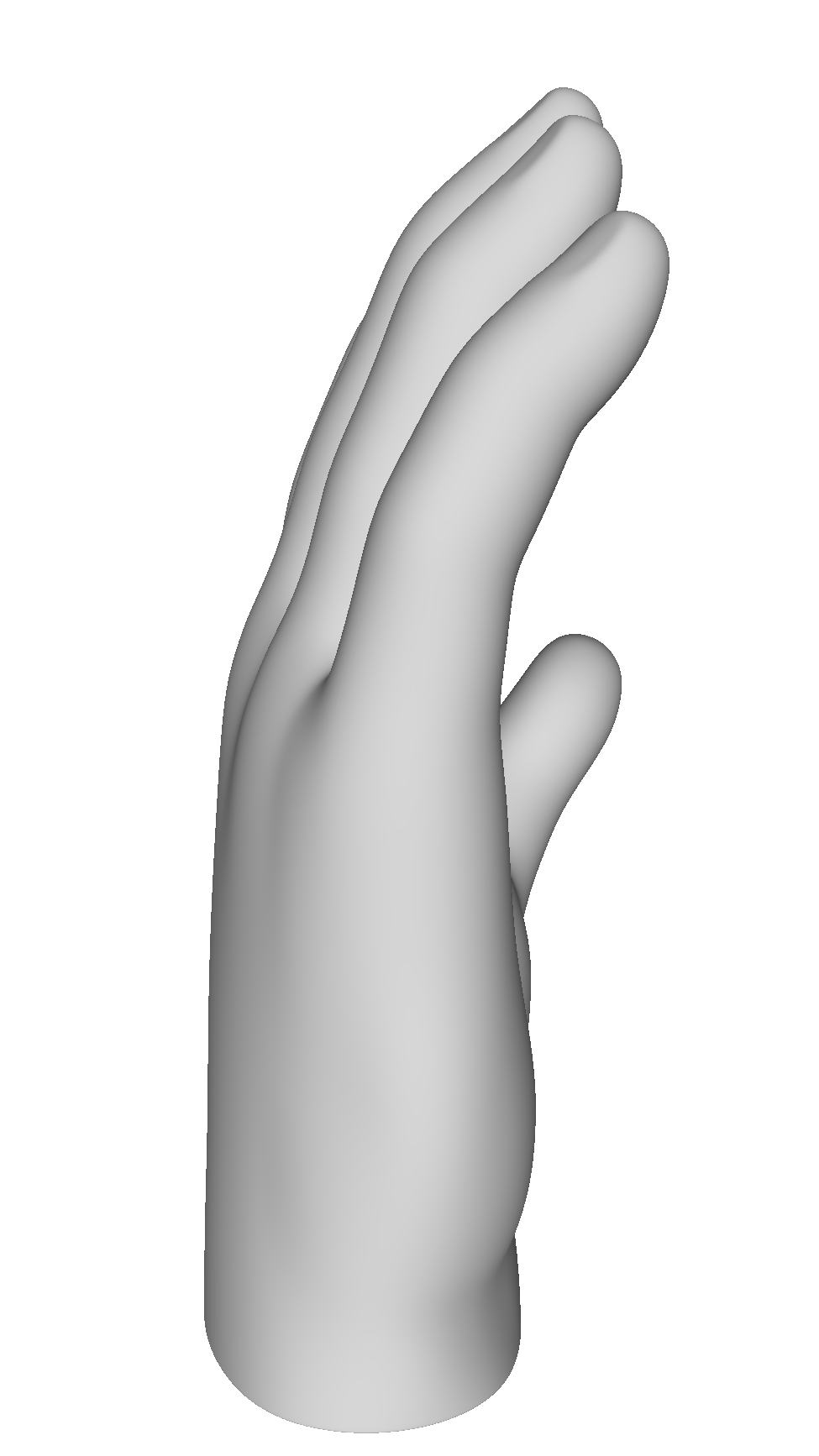}
         \caption{"Perpendicular" hand gesture}
         \label{fig:perp}
     \end{subfigure}
     \begin{subfigure}[b]{0.23\textwidth}
         \centering
         \includegraphics[width=\textwidth]{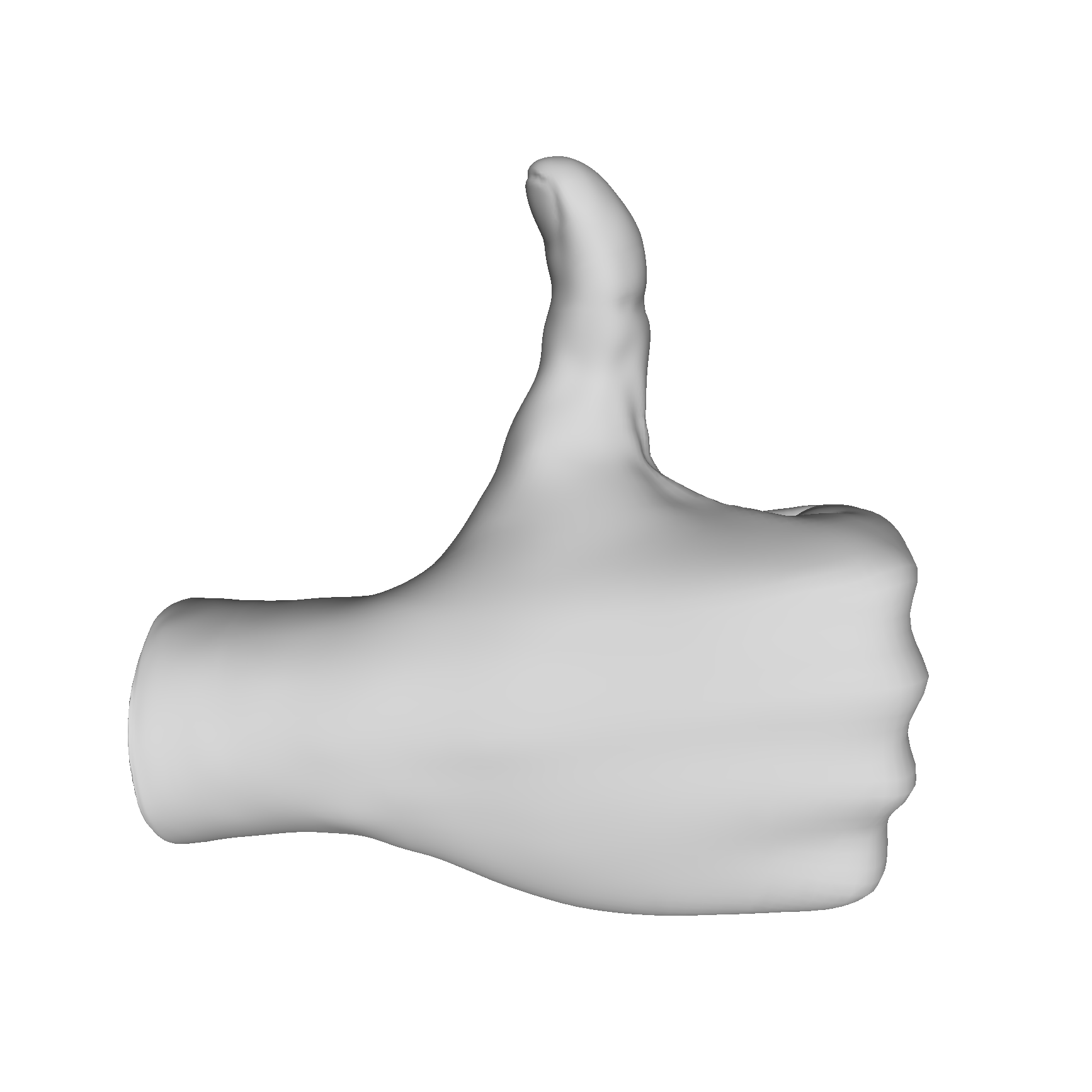}
         \caption{"Thumbs-up" hand gesture}
         \label{fig:tu}
     \end{subfigure}
        \caption{Three static hand gestures from the perspective of the radar.}
        \label{fig:hand_gestures}
\end{figure}

We proceed to test the proposed system using three distinct static hand gestures which we label "palm", "perpendicular", and "thumbs-up," as shown in Fig. \ref{fig:hand_gestures}. For the "palm" gesture, the user places their hand with the palm facing the radar. The "perpendicular" gesture involves the subject's hand oriented perpendicular with the $x'$-$y'$ plane and thumb facing away from the radar. And, the "thumbs-up" gesture requires the subject to face the back of their hand towards the radar with all fingers abducted except the thumb which points upward. 

As mentioned previously, the RCS of the human hand is problematically small in comparison to noise and propagation effects. Comparing the range profiles of the different gestures, the differences are mostly indistinguishable to the human eye. Even though a peak exists in the range FFT at the distance corresponding to the human hand, the features of the gesture reflected back to the radar are not sharply defined and are centered at different places on the human hand. 

\begin{figure}
     \centering
     \begin{subfigure}[b]{0.35\textwidth}
         \centering
         \includegraphics[width=\textwidth]{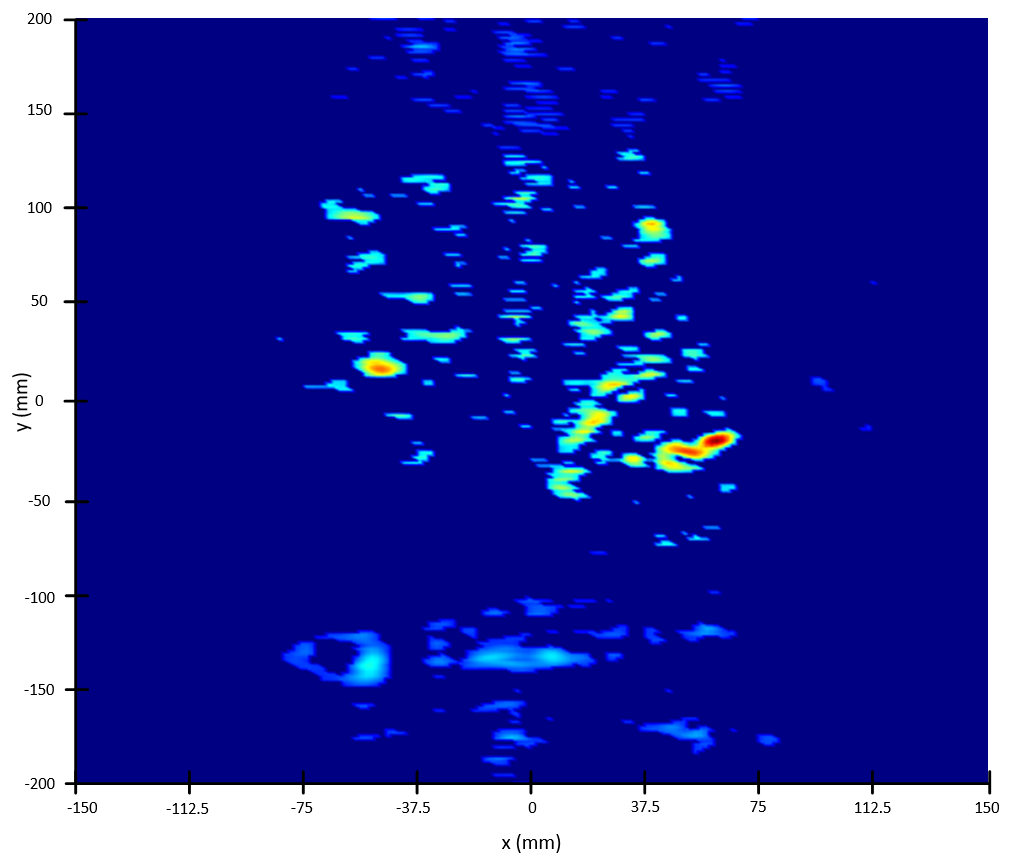}
         \caption{SAR image of a real human hand at $0.25$ m from the radar}
         \label{fig:sar_real_hand}
     \end{subfigure}
     \begin{subfigure}[b]{0.35\textwidth}
         \centering
         \includegraphics[width=\textwidth]{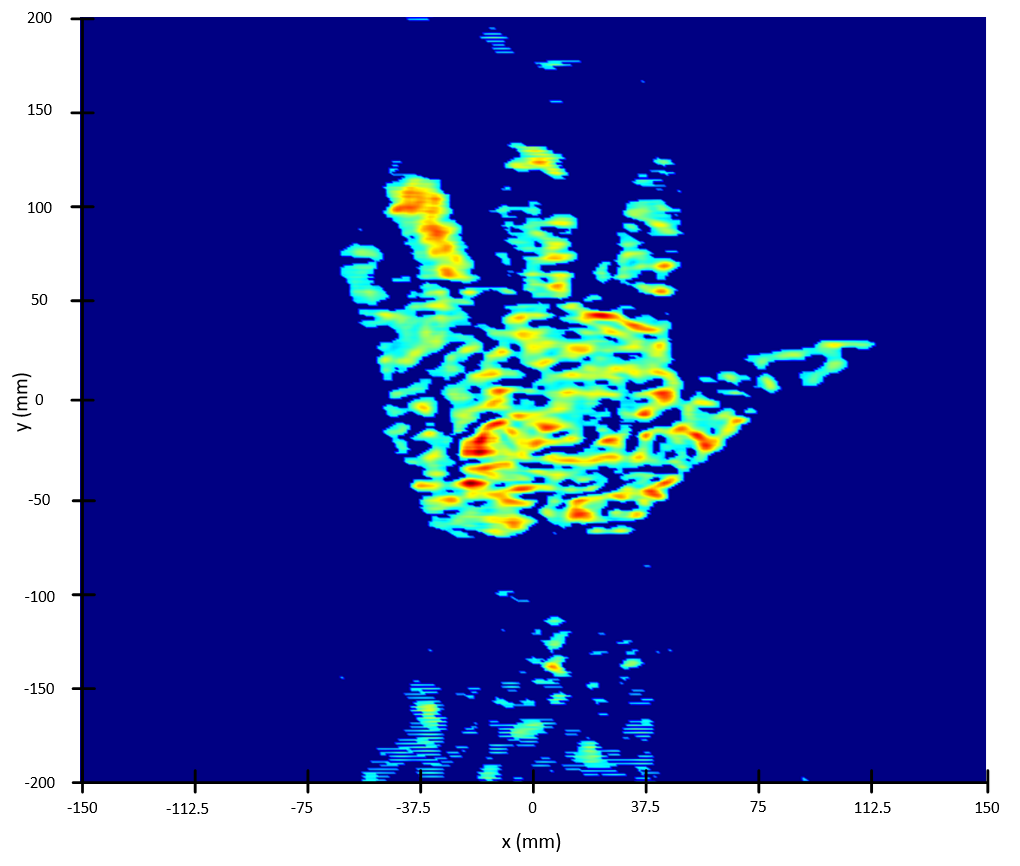}
         \caption{SAR image of an aluminum cutout hand at $0.25$ m from the radar}
         \label{fig:sar_foil_hand}
     \end{subfigure}
        \caption{Comparison of the reconstructed SAR images from the real human hand and the aluminum cutout of the human hand demonstrating the low RCS of the human hand. For consistency, both the human hand and the aluminum cutout are placed on top of a tripod, which contributes the reflections visible beneath the hand.}
        \label{fig:sar_images}
\end{figure}

To demonstrate this phenomenon, a synthetic radar aperture (SAR) approach is temporarily adopted to reconstruct an image of the human hand using the methods described in \cite{yanik2020development,sheen2010near}. It is important to note that the images shown in Fig. \ref{fig:sar_images} are not the data used to train and validate the CNN. These images require all the data from the entire horizontal and vertical scan, which takes approximately $5$ minutes to complete. The data used to train the network are discussed in greater detail later. 

The reconstructed image of the human hand, Fig. \ref{fig:sar_real_hand}, shows a poor image of the hand due to the low RCS. For comparison, a SAR image is also reconstructed using an aluminum cutout in the shape of the hand attached to demonstrate an ideal hand target, Fig. \ref{fig:sar_foil_hand}. For the remainder of this article, we call data collected using aluminum cutouts "sterile" data. This empirical analysis uncovers the innate difficulty in classifying hand gestures from the radar beat signal. Even combining thousands of radar return signals to construct the SAR image, the hand is barely visible and the gesture is difficult to recognize. From these images, we can infer that the features from a human hand contained in a single beat signal reflected are not pronounced and have a quite low magnitude compared to the surroundings, noise, etc. On the contrary, from Fig. \ref{fig:sar_foil_hand}, the aluminum cutout demonstrates a high SNR, meaning the features of the gesture are much more prominent and consistent for each static gesture. The novel data collection strategy proposed in this article consists of capturing data from many perspectives using a 2-D mechanical scanner from both "real" human hands and "sterile" aluminum cutouts, in attempt to improve classification accuracy.

From these observations, we pose several key questions this article aims to answer. Does the radar return signal from an aluminum cutout of a static gesture contain more pronounced, meaningful, and consistent features uniquely describing each gesture compared to the return signal from a human hand? If so, can these "sterile" radar data captured from aluminum cutouts be used to improve the accuracy of a CNN classifier? Specifically, will a training set consisting of both human hand data and "sterile" data provide more easily learnable features to the CNN?

\section{Deep Convolutional Neural Networks}
\label{sec:dcnn}
Before answering those exploring those questions further, we first will overview the proposed classifier. In data-driven detection problems, there are two fundamental steps to constructing a robust classifier: feature extraction and classification. Many methods have been applied to extract features from raw data including handcrafted features \cite{zhang2016dynamic,kim2009human}, linear predictive coding \cite{javier2014application}, empirical mode decomposition \cite{fairchild2014classification}, principal component analysis \cite{kim2014detection}, and more. Similarly, many different classification techniques have been adopted such as k-nearest-neighbors (KNN) \cite{wan2014knn}, support vector machines \cite{zhang2016dynamic,kim2017staticgesture}, dynamic Bayesian networks \cite{pavlovic1999dynamicBayesiannetworks}, etc. 
However, in recent years, the preferred approach for many classification and regression problems is the deep convolutional neural network \cite{kim2016hand}. The concept of deep learning combines feature extraction and classification into a singular step. Now, the feature extraction and classification are simultaneously modeled as a single optimization problem. CNNs have been widely used for image classification and are popularly employed for dynamic gesture recognition on Doppler radar \cite{dekker2017gesture,kim2017application}. Unique from most conventional machine learning algorithms, CNNs adopt a multi-layer approach with interconnected neurons meant to imitate the human brain. Further, due to recent advancements in parallel computing, specifically in graphics processing units (GPUs), training CNNs with complex architectures consisting of many layers has become increasingly feasible. 

The fundamental building blocks of a CNN are convolution layers and nonlinear activation functions. The convolution layers extract features by convolving an array of weights over the input image. The weights are updated every iteration by the back-propagation algorithm used in conjunction with stochastic gradient descent to maximize the classification rate and minimize the loss. After the convolution layer, a nonlinear activation function is applied. Most deep CNNs employ a Rectified Linear Unit (ReLU) defined as $f(x) = \max(0,x)$ over the traditional sigmoid function for improved results \cite{glorot2011ReLU}. By using a nonlinear activation function, the network is able to learn the highly nonlinear complex relationships between the inputs and outputs. 

After convolution and activation, pooling layers are often used to downsample the data by either the average or maximum of a local pool. Convolution, ReLU, and pooling layers are connected to form a complex network of neurons and are finally followed by a fully connected layer, which reduces the dimensionality to the known number of classes, and subsequent general perceptron for classification.

\subsection{Preprocessing of Input Images}
\label{subsec:cnn_input_images:complex_beat_signal}
Input data are gathered from the radar and undergo preprocessing before being used for network training and validation. First, the multistatic-to-monostatic conversion in (\ref{eq:mult-to-mono}) is applied to the complex-valued, MIMO beat signal described in equation (\ref{eq:fmcw_beat_signal_k_multistatic}). Then either the range or range-angle analysis described in Section \ref{subsec:fmcw_range_angle_analysis} is performed. In the next section, we compare the results from using only range analysis to those using range-angle analysis. The complex-valued image is of size $8$x$N_R$, for the range analysis case, or $M_A$x$N_R$, for both range and angle analysis, where $M_A$ is the number of angle FFT bins and $N_R$ is the number of range FFT bins. 

Since minute variations in the hand reflectivity are contained in the phase of the radar beat signal, retaining the amplitude and phase of the complex-valued signal is essential for accurate classification. Some work has been done towards complex-valued implementations of CNNs for radar problems such as SAR image classification \cite{zhang2017complex} and enhanced SAR imaging \cite{gao2018enhanced}, however, we consider an alternative approach to classifying the complex radar return signals. Rather than simply taking the magnitude of each image pixel, the real and imaginary parts of the range or range-angle data sample are layered, forming images of size $8$x$N_R$x$2$ or $M_A$x$N_R$x$2$. Now, inherent relations between the magnitude and phase of the radar data are not lost, improving the classification rate. 

Additionally, most deep CNN implementations employ an input normalization for each channel for numerical robustness. In this case, however, normalizing the real and imaginary layers effectively ruins the phase interdependence. As such, prior to separation into real and imaginary layers, the complex-valued image is normalized to zero-mean and unit variance. Then, no normalization is applied to the real-valued 3-D arrays.

\section{Classification Results}
\label{sec:classification_results}
In this section, we use empirical results to affirmatively answer the questions posed in Section \ref{subsec:challenges} demonstrating for effectiveness of our proposed "sterile" radar data collection techniques for improving static hand gesture classification using mmWave radar. Supplementing training data with synthetically generated data for convolutional neural networks has proven effective for numerous deep learning problems \cite{richardson2016_3D_face_synthetic,wu2019_FaultSeg3D_synthetic,allken2019fish_synthetic,bjorklund2017licenseplate_synthetic}. However to our knowledge, this work is the first to use synthetic "sterile" hands, in the form of aluminum cutouts, captured by mmWave radar, to improve the classification rate of static hand gestures.

\subsection{Training Data}
\label{subsec:training_data}
As discussed in Section \ref{subsec:fmcw_range_angle_analysis}, a 1-D range FFT or 2-D range-angle FFT is applied to each captured beat signal prior to training. For the remainder of this article, we will refer to these datasets as the "range" and "range-angle" datasets, respectively. Each dataset consists of $80000$ samples for each of the three static hand gesture classes of which $40000$ are from human hands and $40000$ are from aluminum cutouts. These samples are rapidly captured at various locations throughout the 2-D $x'$-$y'$ plane by moving the radar mounted to the mechanical scanner. Both the test subjects' hands and the aluminum cutouts vary in size and shape pursuant of improving classifier robustness. Eight participants are selected with varying age, height, weight, hand size and shape, arm length, torso size, and gender. 

To construct the aluminum cutouts, a simple procedure is adopted. Hands of different sizes and shapes are selected from online sources. These shapes are cutout from a thin aluminum sheet and attached to an equivalently shaped cardboard cutout, allowing for a simple training process for sterile data collection. Similarly, eight aluminum cutouts are prepared for each gesture class and used to collect the data. Each aluminum cutout does not directly correspond to test subject hand; rather, the aluminum cutout sizes and shapes are selected independently. For the range data, FFT is performed across the $k$-domain yielding the range profile. The region wherein the hand is expected to be placed is selected at a size of $64$ range bins. Similarly, for the range-angle dataset, the range FFT is performed followed by an angle FFT of size $16$. Once the data are preprocessed, the range dataset images are of size $64$x$8$x$2$ and the range-angle images are of size $64$x$16$x$2$. 

\begin{figure}[h]
    \centering
    \includegraphics[width=0.5\textwidth]{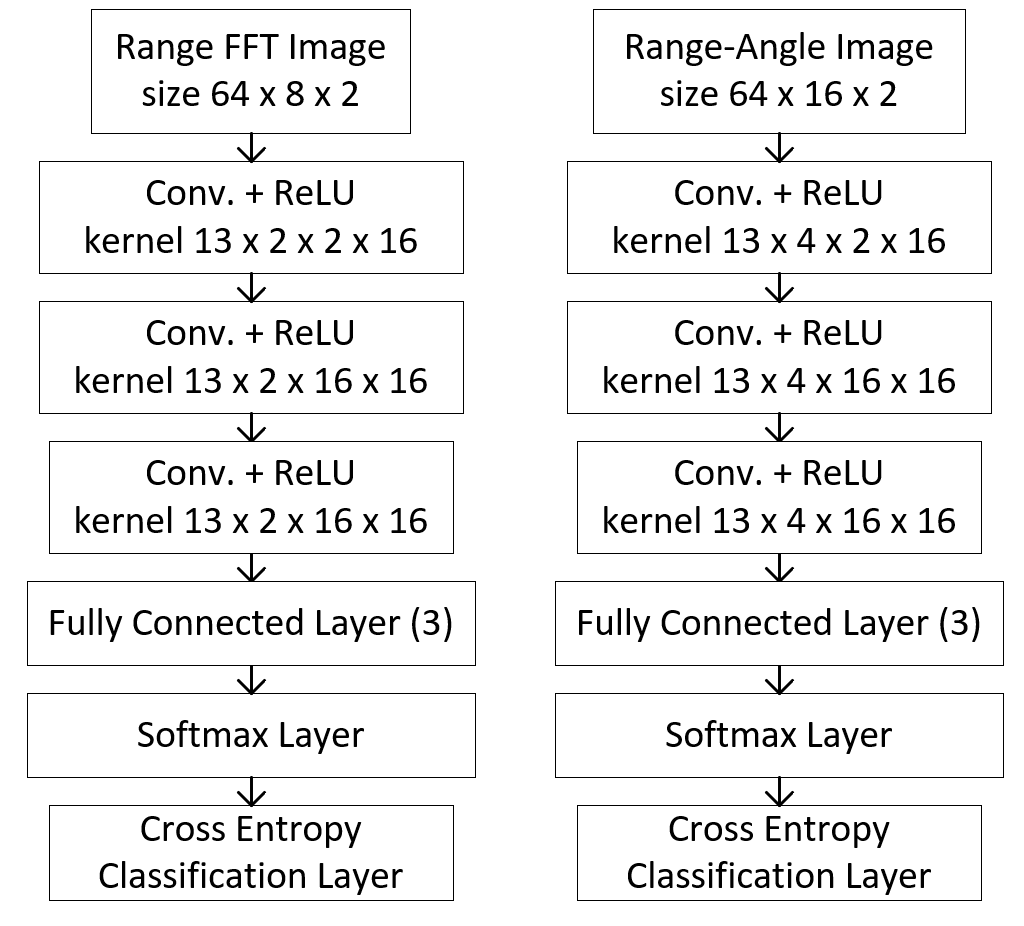}
    \caption{The network architecture for the Range FFT CNN and Range-Angle FFT CNN.}
    \label{fig:cnn_architectures}
\end{figure}

\subsection{Network Architecture and Training}
\label{subsec:network_architecture}
The networks used to classify the hand gesture vary based on the preprocessing applied to the dataset. For the range dataset, convolutional layers with kernel sizes of $13$x$2$ each with $16$ filters are each followed by a Rectified Linear Unit. These are connected in series followed by a fully connected layer with $3$ output neurons, a softmax layer, and a final classification layer using the cross-entropy loss function. The range-angle dataset employs a network with the same architecture changing only the size of the convolutional layers to $13$x$4$ to account for the larger image sizes. Key to both networks are the complex-valued layering and network architectures. Considering the real and imaginary parts of the radar range data as distinct layers of the image allows the network to identify pixel-to-pixel relationships as well as layer-to-layer relationships, which correspond to phase information of each complex-valued pixel. The architectures of both networks are chosen after close inspection of the feature sizes in the observation image domain in both range and channel/angle in addition to extensive testing to optimize real-time implementation efficiency and classification rate. Both network architectures are shown in Fig. \ref{fig:cnn_architectures}. 

For training, each dataset is split into $90\%$ for training and $10\%$ for validation, where all the "sterile" data is contained in the training dataset and the validation dataset consists exclusively of randomly selected real hand gestures. Of the training data, $55.6\%$ are captures of the aluminum cutouts and $44.4\%$ are from real human hands. In this way, the sterile data is used to supplement the training dataset but is not included in the validation dataset. To ensure consistency, we set aside the randomly selected captures of the validation dataset to be reused for each experiment, numbering $8000$ in total.

\subsection{Improved Classification with "Sterile" Data}
\label{subsec:improved_classification_with_sterile_data}
To compare against a control, we first train two networks using only the real human hand data. For these networks, we use the $8000$ set aside captures as the validation dataset, making the split between training to validation $80\%$ to $20\%$. After training each network with only real human hand data, the range CNN and range angle CNN have classification rates of $84.9\%$ and $90.2\%$, respectively. These networks are named "Human Only" in Table \ref{tab:results} since they are trained with only range and range-angle profiles from human hands. 

Next, two new networks are trained using the complete datasets, consisting of real human hand data supplemented by "sterile" data from aluminum cutouts. These networks are dubbed "Combined" since they are trained with both real and "sterile" images. It is important to note that the "Combined" networks are validated with the same validation data as the "Human Only." The only difference is the training dataset used for each network. Once trained, the networks corroborate our hypotheses on training with "sterile" data as the classification rates improve to $93.1\%$ and $95.4\%$ for the range and range-angle datasets, respectively. 

\begin{table}[h]
    \centering
    \large
    \begin{tabular}{c|c c}
         & Human Only & Combined \\
         \hline
         Range & 84.9\% & 93.1\% \\
         Range-Angle & 90.2\% & 95.4\%
    \end{tabular}
    \caption{Comparison of classification rate between networks trained with only human hand data (Human Only) and networks trained using sterile data to supplement the real human hand data (Combined).}
    \label{tab:results}
\end{table}

These results promote an affirmative answer to the questions posed earlier. Namely, the network trained on the dataset consisting of both human hand data and "sterile" data is able to learn meaningful features in classifying human hand data more effectively than the network trained using only non-"sterile" data. Further, our results explicitly demonstrate an increase in classification accuracy by employing the "sterile" radar data collection scheme proposed in this article. 

Finally, compared to past work in the literature, our proposed method improves upon gesture recognition by using sterile data while offering a solution to the difficult classification problem of static gestures under three-dimensional spatial translation. Kim \textit{et al.} \cite{kim2017staticgesture} employ a time-domain gesture recognition approach on ultra-wideband impulse-radio radar. \cite{kim2017staticgesture} considers two scenarios separately: (1) six gestures using human hands $15$ cm away from the transceiver and (2) three plaster model gestures rotated at $10^\circ$ increments. For scenario (1), the hand is kept at a constant position for all captures. Both training and testing are performed using human hand data resulting in a classification rate of $91\%$ using a CNN. In scenario (2), plaster models of each gesture are captured from different perspectives by rotating the plaster model. For this scenario, Kim \textit{et. al.} record classification accuracies of above $90\%$ for three gestures, validating using data from the plaster model. Comparatively, our method yields a more robust classifier by including both real human hand reflections and "sterile" reflections in the training processes and validating with only human hand data. Rather than creating two distinct classifiers for human and sterile data separately, as discussed in \cite{kim2017staticgesture}, the technique proposed in this paper unites human and sterile data to construct a robust classifier. Further, our approach investigates more diverse scenarios by capturing data from multiple test subjects at many locations. 

Extensive work has been done towards dynamic gesture recognition on mmWave radar, Doppler radar, and IR-UWB sensors \cite{kim2017application,dekker2017gesture,suh2018_24GHz_gesture,kim2016hand,zhang2016dynamic,leem2020detecting,park2016_ir_uwb_gesture}; however, this is an entirely separate problem from the problem addressed in this paper as the dynamic gesture case considers only motion. This reduces the dimensionality of the classification to temporal motion features, whereas static gesture recognition on mmWave radar involves classification of a three-dimensional structure using lower dimensional data, as discussed in Section \ref{subsec:distributed_target}. Thus, our model is trained for the more difficult problem of static gesture classification under spatial translation and demonstrates superior classification accuracy to prior static gesture classification work \cite{kim2017staticgesture}.

\section{Conclusion}
\label{sec:conclusion}
In this article, we investigated novel data collection and training techniques for improving the classification of static hand gestures using mmWave FMCW radar and convolutional neural networks. A novel data collection technique for static hand gestures is proposed consisting of a radar mounted on a two-dimensional mechanical scanner allowing the efficient collection of large, diverse radar datasets. Then, we examine the innate challenges of static hand gesture recognition and observe the low radar cross-section of static hand gestures, compared to an ideal aluminum cutout of the same shape. From this observation, we hypothesize that if a convolutional neural network is trained with data from both a real human hand and "sterile" aluminum cutout, the resulting network will outperform a network trained on human data alone because the features unique to each static gesture are more pronounced in the "sterile" data and will be thus easier for the CNN to learn. From this hypothesis, we extend the data collection approach to capture radar range and range-angle profiles of both human hands and aluminum cutouts for use in CNN training.

Three static (non-moving) hand gestures are considered applying both range and range-angle preprocessing. Using deep CNNs and data from human hands only, the classification accuracies for range and range-angle preprocessed data are $85\%$ and $90\%$ respectively. However, using the same data for validation and only changing the training data as described by our hypothesis, the classification rates improve, respectively, to $93\%$ and $95\%$. The increase in accuracy demonstrates the improvement introduced by the novel "sterile" radar data technique never before examined in the literature. Further, the model developed in this article outperforms prior work on static gesture recognition on a more challenging classification problem \cite{kim2017staticgesture}. Since the CNN model relies on the availability of a large amount of meaningful data, such "sterile" and synthetic data acquisition and generation techniques are likely to increase as they have been proven suitable for many classification and regression problems. In future work, we plan to extend this premise to capture more "sterile" static gestures and even "sterile" dynamic gestures to improve CNN accuracy and robustness.

\printbibliography

\end{document}